\documentclass[a4paper]{article}

\usepackage[pages=all, color=black, position={current page.south}, placement=bottom, scale=1, opacity=1, vshift=5mm]{background}

\usepackage[margin=1in]{geometry} 

\usepackage{amsmath}
\usepackage{amsfonts}
\usepackage{lipsum}
\usepackage{algorithm2e}
\usepackage{graphicx}
\usepackage{subcaption}
\usepackage{float}
\usepackage{pifont}
\usepackage{bbding}
\usepackage{epstopdf}

\usepackage{subcaption}

\usepackage[utf8]{inputenc}
\usepackage{hyperref}
\hypersetup{
	unicode,
	pdfauthor={Vahid Tavakol Aghaei},
	pdftitle={Deep Reinforcement Learning Off-policy Algorithms and Benchmark for Solving Various Robotic Manipulator Tasks},
	pdfsubject={A simple article template},
	pdfkeywords={},
	pdfproducer={LaTeX},
	pdfcreator={pdflatex}
}

\usepackage[authoryear,round]{natbib}

\usepackage{graphicx, color}
\graphicspath{{figures/}}

\usepackage{algorithm, algpseudocode} 
\usepackage{mathrsfs} 

\usepackage{lipsum}

\title{Off-Policy Deep Reinforcement Learning Algorithms for Handling Various Robotic Manipulator Tasks}
\author{Altun Rzayev$^1$ \and Vahid Tavakol Aghaei $^2$\thanks{Corresponding author} }

\date{
	$^1$ Ozyegin University, Biomechatronics Laboratory, Istanbul, Turkey\\ 
	\texttt{altunrzayev@gmail.com}\\[2ex]%
	$^2$Istinye University, Faculty of Engineering and Natural Sciences, Electrical and Electronics Engineering, Istanbul, Turkey\\ 
	\texttt{vahid.aghaei@istinye.edu.tr}
}
\begin{document}
	\maketitle
	
\begin{abstract} 
In order to avoid conventional controlling methods which created obstacles due to the complexity of systems and intense demand on data density, developing modern and more efficient control methods are required. In this way, reinforcement learning off-policy and model-free algorithms help to avoid working with complex models. In terms of speed and accuracy, they become prominent methods because the algorithms use their past experience to learn the optimal policies. In this study, three reinforcement learning algorithms; DDPG, TD3 and SAC have been used to train Fetch robotic manipulator for four different tasks in MuJoCo simulation environment. All of these algorithms are off-policy and able to achieve their desired target by optimizing both policy and value functions. In the current study, the efficiency and the speed of these three algorithms are analyzed in a  controlled environment. 
\end{abstract}

\maketitle
\section{Introduction} \label{sec:introduction}
More and more uses for robots are being found, including autonomous vehicles \citep{driving}, industry-based robots \citep{peiman22}, and humanoid robots \citep{humanoid2020}, where the operational landscape is not static, preset and known. Manufacturing  robots, in particular, need to be able to perform intelligent actions in highly variable settings. The primary methods of functioning for conventional automated robots are learned imitation and programming skills. Automated  robots in the industry have low responsiveness once they are put in a situation with unknown variables. Robots typically need the capacity to physically interact with items because they are deployed in unpredictable environments. Positioning, dynamic motion plannings, and grasping are typically at the forefront of such tasks \citep{obstacleavoid22}.  As a result, it is no longer feasible to program every conceivable set of circumstances in advance.  With the need to perceive and assess the operational environment and make judgments for the proper implementation of the tasks taking into account the safety, productivity, and manufacturing challenges, intelligence must be built into the robotic system. Thus, Artificial Intelligence (AI) and Machine Learning (ML) techniques find a vast amount of application in the field of robotics to realize such lofty objectives \citep{Shahid2022roboticRL}.

Reinforcement learning (RL) \citep{BelAbdKliParPet21Book} is a sub-field of ML and has gained huge interest, especially in control and robotics areas, due to its capability of handling complex high-dimensional environments and its successful applications can be found in the works done by \cite{peters2022RL, TAVAKOLAGHAEI2021, emreugur20}. RL continuously interacts with its surrounding environment and gathers data trajectories to decide about the future actions of the controller and its working principle depends on a reward-penalty logic. The long-term goal of RL is to find an optimal control policy that maximizes the expectation of a given objective function. RL approaches have been effectively implemented in a broad range of exciting applications in the field of robotics, such as table tennis game \citep{tennis20IROS, peterstennis22}, acrobatic maneuvers \citep{flip2021}, solving Rubik' cube \citep{rubikcube}, and pancake flipping \citep{pancake22}.

Basically, RL is categorized in two groups: model-based where the controller (agent) needs to learn the environment's model, and then it predicts the future data \citep{MBRL22survey}. Some researchers believe that model-based RL could benefit more from sample efficiency than model-free algorithms. While the former is promising, but it has obstacles that prevent it from being extensively used in the real life situations, such as the difficulty of systematically and effectively developing a precise model from pure data acquired.
In order to acquire low-dimensional implicit state and action interpretations from large-dimensional data sets, several publications \citep{Hafner1, Hafner2} employ representational learning techniques. However, there is a chance that the trained models will not agree with the underlying dynamical mechanisms, and their efficiency can also decrease dramatically if we move beyond the range of the training sets. However, with the recent developments in visualization and classical mechanics based on simulation settings \citep{pancake22}, robots are now able to efficiently manipulate based on predetermined modeling framework via graphical algorithms.

The model-free concept is an alternative that seeks to discover the optimum solution without first estimating the model. The model-free technique has the potential to be less space-intensive than its model-based analogue because there is no longer any need to save a detailed explanation of the system \citep{modelfree21_1}.

One disadvantage for model-based RL approaches is that they sometimes suffer from learning complex models accurately as well as the increased training time. In order to boost the training procedure and obtain the optimal result quickly, off-policy model-free RL algorithms are preferable \cite{c20}. They are advantageous since they use batch learning from their past experiences and achieve significant success during training. The agent can acquire knowledge about additional policies that are distinct from the one currently being carried out by engaging in off-policy learning. \cite{offpolicy18imani} provides a comprehensive analysis of the benefits of off-policy optimization. Leveraging the function approximation potential of deep neural networks has led to the widespread use of RL. Deep reinforcement learning is a set of methods that use neural networks to learn interpretations from large feature dimensions, allowing for the learning of control in a holistic fashion. Among those algorithms, Deep Deterministic Policy Gradient (DDPG) \cite{ddpg15}, Twin Delayed Deep Deterministic Policy Gradient (TD3) \cite{twin18} and Soft Actor-Critic (SAC) \cite{SAC18}, are common examples of off-policy algorithms. All the above algorithms optimize both the policy and the value functions and maximize the cumulative reward to converge to the optimal result.\\

DDPG is the most used off-policy algorithm for the systems with continuous state-action spaces. Apart from robotic control, DDPG is also used for energy efficiency of wind turbines \cite{DDPGwind}, optimization of unmanned aerial vehicles \cite{UAV21}, and traffic management \cite{trafficDDPG} due to its fast convergence. \cite{TuyenC17} controlled the bicycle without human interaction, which is one of the most challenging tasks in control area using DDPG. Nevertheless, DDPG, which even exhibits a high performance on humanoid robots \citep{kumar18} sometimes may fail to conquer during training process due to the overestimation of the Q-value function. On the other side, TD3 algorithm is presented to bring a solution for this overestimation issue by using two Q-functions. TD3 looks like DDPG in terms of algorithmic structure and is frequently used for Energy Systems\citep{woo20} and Robotic Control \citep{Dankwa19}. Unlike DDPG and SAC, TD3 updates the policy network less frequently.\\
SAC is also another off-policy algorithm which maximizes the expected return and the entropy. The structure of the algorithm is different from DDPG and TD3. This entropy regularization method is long ago used for optimal control \citep{RawlikTV12, Toussaint09, Todorov08}  and inverse reinforcement learning problems \citep{ZiebartMBD08} and achieved high performance. This algorithm is preferred due to its exploration method and the entropy regularization. \\
Similar work has recently been done for robotic manipulations by \cite{Aumjaud2020}, in which the Advantage Actor-Critic (A2C) \citep{Mniha2c16}, Actor-Critic using Kronecker-Factored Trust Region (ACKTR) \citep{wu17}, DDPG, TD3, PPO, SAC, and TRPO algorithms are applied and compared for target reaching tasks in \emph{Pybullet} simulation environment and on a physical setup. In this study, the DDPG, TD3, and SAC algorithms are applied to a 7-DoF robotic manipulator to perform four different tasks in the \emph{MuJoCo} environment: Reach, Pick and Place, Slide, and Push \citep{c18}.
This study exposes the algorithms to a more broad comparison by evaluating them on tasks with varying levels of difficulty in order to produce significantly more dependable results.

The following contributions can be covered under the scope of this study:
\begin{itemize}
	
	\item Model-free off-policy RL algorithms (DDPG, SAC and TD3) are analyzed and applied to a Fetch robot manipulator in the MuJoco simulation environment to learn different robot manipulation and control tasks and their learning performance is compared. In all cases, the controller agents are acquired through reinforcement learning in a continuous space defined by a custom reward function. The simulation of corresponding tasks can be reached via: \url{https://www.youtube.com/watch?v=ce-PuCThzBw&list=PLLNrNJhfBBw1SgwuUE9H-mgLtmnzt4fEx}
	\item To choose the most suitable model-free algorithm for each specific task from DDPG, SAC and TD3, a comparison has been made in terms of success rate, time complexity and efficiency.x
\end{itemize}	
This article is organized as following: After a literature background in Section \ref{sec:introduction}, deep RL algorithms applied to the Fetch robotic system are described in Section \ref{sec: algorithms}. Section \ref{sec: environment} explains the Mujoco simulation environment and the related robotic manipulator. Experimental results are given in Section \ref{sec: results}, and finally the concluding discussions regarding the experiments are provided in Section \ref{sec: conclusions}.

\section{RL: A General Overview}\label{sec: algorithms}
\subsection{RL structure}
Machine learning (ML) is the process by which a computer is taught to improve its overall quality on a certain function by using the information it has already amassed. Supervised, unsupervised, and reinforcement learning are the three main categories into which ML algorithms fall. Unlike unsupervised learning, which makes the use of methods like cluster analysis on raw data sources, supervisory techniques are founded on inference theory, with the model being trained on labeled data to carry out regression or classification . However, according to the RL concept, an autonomous operator can improve its performance on a task through observation and experimentation. On the other side, agent, is anything capable of gathering information about its surroundings via sensory devices and then responding to what it finds.
RL structures, for formalizing the sequential inference events, can be identified as Markov decision Processes (MDP) for stochastic control problems. MDP is a method of decision-making that bases its considerations primarily on the latest recent policy and behavior, rather than the complete decision-making record. The decision-making dilemma is classified as a partially-observable MDP because, more significantly, in many realistic areas of application, an actor cannot see all elements of the environmental state . The objective of reinforcement learning is to locate a policy that, when applied to histories in the state space, would result in the maximization of the estimated summation of (discounted) rewards. MDP can be thought as a set ($\mathcal{S},\mathcal{A},\mathcal{P},\mathcal{R}$), where $\mathcal{S}\subset\mathbb{R}^{d_s}$ constitutes the continuous state, $\mathcal{A}\subset \mathbb{R}^{d_a}$ continuous action space, $\mathcal{P}$ is the state transition function, and $\mathcal{R}\subset\mathbb{R}$ determines a real-valued reward function to assess the quality of the state transitions from a present state $s$ to a new state $s'$. A stochastic policy $\pi_{\theta}(a|s)$ dependent to some parameters $\theta$ aims at mapping the states of the system to the action space by maximizing the average of the obtained rewards in the long run.

We focus on recurrent encounters and gather state and action combinations using the notation $x_{t} := (s_{t}, a_{t})$  in order to create governing patterns over the course of a certain timescale. The series $\{ x_{t}: t \geq 1 \}$ is an example of a Markov process that operates on the space $\mathcal{X} = \mathcal{S} \times \mathcal{A}$. The transition function for this sequence may be expressed as

\begin{equation*} 
	f_{\theta}(x_{t+1} | x_{t}) := g(s_{t+1} | s_{t}, a_{t}) \, h_{\theta}(a_{t} | s_{t}).
\end{equation*}

The distribution for a specific route, denoted by $x_{1:n}$, that leads from a starting state to a final time denoted by $n$ can be expressed mathematically as follows:
\begin{equation*} 
	p_{\theta}(x_{1:n}) :=f_{\theta}(x_{1}) \prod_{t =1}^{n-1} f_{\theta}(x_{t+1} | x_{t}),
\end{equation*}
where $f(x_{1}) = \eta(s_{1}) h_{\theta}(a_{1} | s_{1}) $ is the initial distribution for $x_{1}$.

A presumed return component is allocated, and its value might be the whole or discounted sum of immediate rewards. This function is utilized in the long-term to assess the efficiency of a path that was obtained. The discounted form, in which the reward at every time interval is calculated by a rate of discount denoted by $\gamma \in (0,1]$, is the one that will be considered in this article. 
The discount factor is what decides how an agent should prioritize the rewards that will come in the future. Agents who have a small $\gamma$ are more probable to have a short-term focus and try to maximize their profits in the short future. On the other hand, agents whose values are high are more likely to have a long-term perspective and try to maximize their gains over the course of a longer time frame.

The following formula can be used to formulate a specific path's return:
\begin{equation*} 
	R(x_{1:n}) := \sum_{t = 1}^{n-1}\gamma^{t-1}r(a_{t}, s_{t}, s_{t+1}).
\end{equation*} 

The primary objective of RL policy searching is to investigate the policy regimes in order to determine the value of the policy variable $\theta$ that optimizes a cost $J(\theta)$ that can be defined as an expected value over the total rewards.
\begin{equation*}
	\theta^{*} = \arg \max_{\theta \in \Theta} J(\theta).
\end{equation*}

\begin{equation} \label{eq: J theta}
J(\theta) = \mathbb{E}_{\theta}[R(x_{1:n})] = \int p_{\theta}(x_{1:n}) R(x_{1:n})dx_{1:n}.
\end{equation}

The integral that is engaged in the formulation of this expectation necessitates it being challenging to evaluate it since it is irresolvable. In most cases, this problem emerges as a result of the undetermined sampling of the histories or the construction of the rewards. Several alternative approaches based on statistical approaches have been used in an effort to locate $\theta^{\ast}$. Among such efforts, policy gradient is noted for being one of the most prominent options; for an early diligent work, refer to the citation\cite{petersPG2006} which is on the basis of usual gradient updates with a learning rate of $\beta$ as:

\begin{equation*} 
	\theta^{(k+1)} = \theta^{(k)} + \beta \nabla_{\theta} J(\theta^{(k)}).
\end{equation*}

However, the gradient is computed by a Monte Carlo method.
\begin{equation*}
	\nabla_{\theta} J(\theta) \approx \frac{1}{N} \sum_{i = 1}^{N} \left[ \sum_{t = 1}^{n}\nabla_{\theta} \log h_{\theta}(a_{t}^{(i)} | s_{t}^{(i)}) \right] R(x_{1:n}^{(i)}),\quad \text{where} \quad  x^{(i)}_{1:n} \sim p_{\theta}(x_{1:n}), \quad i = 1, \ldots, N.
\end{equation*} \\
Almost all practical applications use action domains that are continuous. Deterministic policy gradient (DPG) techniques permit RL to be used in contexts where actions can be performed continuously. Policy gradients with a deterministic structure can be expressed for MDPs satisfying certain requirements by adopting a model-free formulation that tracks the value function's gradient \citep{Silverconf}. This means that in scenarios with larger action spaces, DPG requires fewer observations because it just integrates over the state space, as opposed to the state and action spaces as in probabilistic policy gradients.

Compared to the policy based RL algorithms, on the other side, there exisit the value-based ones such as Q-learning. In value-based methods, the agent learns estimates of the  individual state-action pairs $Q(s,a)$. If enough samples are collected for each state-action pair, Q-learning will learn (near) optimal state-action values. Once a Q-learning agent has converged to the optimal $Q$ values for an MDP and made greedy action choices afterwards, it will get the same expectation of the discounted rewards' summation as estimated by the value function. The updates for each state-action's $Q$ values are done according to:
\begin{equation}
Q(s,a) \leftarrow Q(s,a) + \alpha \left[ r + \gamma \max_{a'\in A} Q(s',a') - Q(s,a) \right]
\end{equation}

To combine the benefits of the policy-based and value-based RL algorithms, the actor-critic methods are proposed. The 'actor' refers to the policy structure that makes decision about which actions to take. In this case, the critic is the estimated value function, which provides feedback on the actor's performance. Every time an action is chosen, the critic assesses the new state to determine if the outcome was satisfactory or not. In this case, learning procedure demands the gradient calculations for both networks \citep{survey2022deep}. To extend the application of $Q$ functions from discrete state-spaces to high-dimensional continuous ones, Deep Q-Networks are introduced where they include deep neural networks for approximating the $Q$ values in complex domains. The technique of experience replay is utilized by DQN in order to both improve the overall efficiency of the collected samples and to cut the correlation that exists between iterative data samples \citep{mnih2015DQN}. One of the successful actor-critic methods for high-dimensional state-spaces, is the improved and extended versions of DQN and DPG, called Deep Deterministic Policy Gradient (DDPG). In this study we will apply DDPG, Twin Delayed DDPG, and Soft Actor Critic (SAC) algorithms to our experiments in the MuJoCo environment and compare the success rate of these algorithms over the different robotic tasks. In the following section, we will have a brief overview to these algorithms.
\subsection{Adopted RL Algorithms}
In this part, we briefly discuss the implemented off-policy algorithms that we have selected for our robotic manipulator.

\subsubsection{Deep Deterministic Policy Gradient (DDPG)}
DDPG with a continuous action domain is a model-free reinforcement learning algorithm. It is an extension of the DQN (Deep Q-Network) algorithm, which is designed for discrete action spaces. DDPG uses two neural networks, called the actor and the critic, to learn a policy and a value function, respectively. The actor network is used to predict the best action to take in a given state, while the critic network is used to evaluate the quality of the actor's chosen action. The two networks are trained using a combination of supervised learning and reinforcement learning techniques. DDPG is well-suited for tasks involving continuous control, such as robotic manipulation tasks. It has been used to solve a variety of challenging problems in this domain, including reaching, picking, placing, and sliding objects.

DDPG is an RL algorithm that concurrently learns the optimal policy and the Q-function. Q-function in DDPG is being learned using Bellman equation and off-policy memory, afterwards the learned Q-function is adopted to learn the policy. The method seeks to determine the optimal action value $a^{*}$ for each given state in a continuous control environment, in a manner similar to the Q-learning method. The learning of Q-function, which uses the Bellman equation as provided in Eq. \eqref{eq: belmann}, is the first mathematical aspect of DDPG.
\begin{equation}\label{eq: belmann}
Q^*(s,a) = \mathbb{E}_{s'\sim P}\left[ r(s,a) + \gamma \max_{a'} Q^* (s', a')\right]
\end{equation}

Here,  $s'\sim \mathcal{P}$ means sampling the next state $s'$ from target transition function $\mathcal{P}(s,a)$. The Bellman equation's role here is to approximate optimal Q-function. This Q-function can be approximated by a set of artificial neural networks $Q_\phi(s,a)$, where $\phi$ is the parameters of the network. The working principle of DDPG is the minimization of the mean-squared Bellman error (MSBE) function defined in Eq. \eqref{eq: MSBE} for a set of data trajectories $\mathcal{D}(s,s,r,s')$:
\begin{equation}\label{eq: MSBE}
\begin{split}
\mathcal{L}(\phi, D) = &\mathbb{E}_{\mathcal{D}}\Bigg[\bigg(Q_\phi(s,a)- \\ 	 &\Big(r+\gamma(1-d)\max_{a'} Q_\phi(s',a')\Big)\bigg)^2 \Bigg]
\end{split}
\end{equation}
Here $(s,s')\in\mathcal{S}$, $(a,a')\in\mathcal{A}$, and $r\in\mathcal{R}$ with a learning rate $\gamma \in (0,1]$ and $d$ signifying that whether a terminal state is met or not.\\
Noise should be added to the policy actions during the training in order to properly explore the policy space. Normally,\cite{ddpg15} suggests utilizing \emph{Ornstein-Uhlenbeck} (OU) noise for DDPG, however Gaussian noise is used in this study because recent studies have shown that OU noise has no unique efficiency over the training data \cite{c17}.
\subsubsection{Twin Delayed Deep Deterministic Policy Gradient (TD3)}
TD3, or Twin Delayed DDPG, is a model-free off-policy reinforcement learning algorithm for continuous action spaces. It is an extension of the DDPG algorithm, which is designed to address some of the limitations of DDPG. TD3 uses two actor networks and two critic networks, which are trained to work together to maximize the agent's performance. The two actor networks are used to generate two different action outputs for each state, which are then evaluated by the two critic networks. This allows TD3 to better capture the uncertainty in the environment and reduce the overestimation of value estimates that can occur in DDPG. Like DDPG, TD3 is well-suited for tasks involving continuous control, such as robotic manipulation tasks. It has been shown to perform well on a variety of challenging problems in this domain.

While the DDPG algorithm benefits from a high efficiency on performance, it is fragile with respect to hyper-parameters and substantially overestimates the Q-values, causing the policy to worsen.

The TD3 algorithm tackles this problem by incorporating three new approaches into the DDPG algorithm. To begin, unlike DDPG, TD3 uses two Q-functions $Q_{\phi_1}(s,a), Q_{\phi_2}(s,a)$ and selects the smallest Q-value according to the Bellman error loss function given in Eq. \eqref{eq: MSBE} and then the policy is learned as
\[
\max\limits_{\theta}\mathbb{E}\Big[Q_{\phi_1}\Big(s,\pi_\theta(a|s)\Big)\Big]
\]
As a result, the new algorithm is dubbed "twin." In the second scenario, the target network and policy are modified less frequently. \cite{c5} state that one policy update for TD3 is advised per every two Q-function updates. The final option is to smooth the target policy, which involves adding noise to the target action in order to make the policy more complicated when dealing with Q-function inaccuracies. Compared to the usual DDPG algorithm, all of these innovative ways enhanced the efficiency of the complicated systems.

\subsubsection{Soft Actor Critic (SAC)}
SAC is an RL off-policy technique that optimizes a stochastic policy to learn an action, combining stochastic policy optimization and DDPG approaches. Although it has no relation to the TD3 method, it uses a double clipped Q-function and aims for policy smoothing due to the policy's intrinsic stochasticity. The entropy regularization feature is the most significant aspect of the algorithm, which is employed for continuous action space. During policy training, the expected return and entropy are maximized in this algorithm. The use of this strategy eliminates improper convergence.\\
The term entropy which evaluates the randomness of a variable $x$ using its density function $P$, can be defined as:
\begin{equation}\label{eq: entropy}
H(P) = \mathbb{E}_{x \sim P}[-\log P(x)]
\end{equation}
Taking into account the effect of the entropy the regularized RL problem can be modified as:
\begin{equation}\label{eq: RL_entropy}
\pi^{*}_\theta = \arg \max \limits_{\pi} \mathbb{E}\Bigg[\sum_{t=0}^{\infty}\gamma^t R(s_t,a_t) + \alpha H\Big( \pi(a_t|s_t)\Big)\Bigg]
\end{equation}
Besides policy $\pi_\theta$, two Q-functions $Q_{\phi_1}, Q_{\phi_2}$ are being optimized by SAC algorithm. For this, the Bellman equation can be reformulated as: 
\begin{equation}
\begin{split}
Q^\pi(s,a) = &\mathbb{E}_{s' \sim P, a' \sim \pi}[R(s,a,s')+ \\
&\gamma(Q^\pi(s',a')-\alpha \log \pi(a'|s'))]
\end{split}
\end{equation}
where, $\alpha > 0$ can be either a fixed or varying coefficient where for this article a fixed version is used. Similar to TD3, SAC algorithm computes MSBE function given in Eq. \eqref{eq: MSBE} for $Q_{\phi_1}$ and $Q_{\phi_2}$ and then opts for the minimum Q-function.

\section{Environment}\label{sec: environment}
Fetch robotic manipulator is used to test the algorithms in the simulated environment whose job is to push or pull a black cube toward a red dot located on a flat surface or in midair. The gripper position is controlled in three dimensions in both cases, while the gripper's opening and closing are handled in the fourth dimension. The states of the system include the linear and angular velocities of the gripper and its positions in the Cartesian space. However, in the Push task, the gripper is always set to be closed, requiring the agent to push the cube on the surface, requiring the agent to learn the physical properties of the cube and surface (in this case, friction).

For the simulations, MuJoCo simulator is chosen which is developed by \citep{c18}, and is used for Multi-joint dynamics with contact. Fetch robotic manipulator has 7-DoF arm and two-fingered parallel gripper. The Fetch robot has already been used by different studies in the domain of deep RL for different tasks \citep{Nguyan18Fetch,8463162,8968488}. Four different tasks can be performed by using this environment and the description of these tasks are given below. 

\begin{itemize}
	\item \textit{Slide}: The manipulator should strike the rubber across the table to slide it to the desired position.
	\item \textit{Pick and Place}: The manipulator should pick up the box from the table and move it to the given position.
	\item \textit{Push}: The manipulator should move the box to the desired position by repulsing it.
	\item \textit{Reach}: The manipulator should move its end effector to the desired positions. \\
\end{itemize}

\begin{figure}[thpb]
	\centering
	\framebox{\includegraphics[scale=0.67]{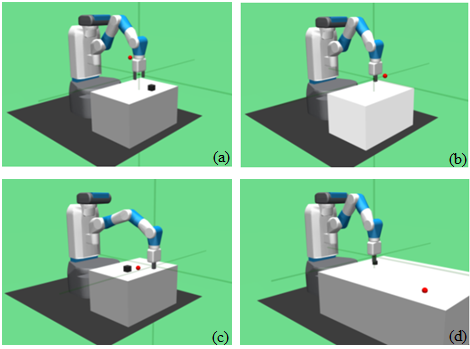}}
	\caption{(a) Pick and places, (b) Reach, (c) Push and (d) Slide from MuJoCo environment}
	\label{figure_envy}
\end{figure}

In this simulation environment, the states of the system consist of the Cartesian positions, the angles and the velocities of all joints. Additionally, the information of achieved and desired goals is obtained from the simulation. The goals of the system are described as desired positions. Sparse and binary rewards are used by the simulator as a default which is -1 if the desired was not achieved and 0 for if it was achieved. The initial position of the gripper is always fixed for the tasks and object location is randomly placed on the table at every episode.

\section{Simulation Studies}\label{sec: results}
\begin{figure}[thpb]
	\centering
	\framebox{\includegraphics[scale=0.57]{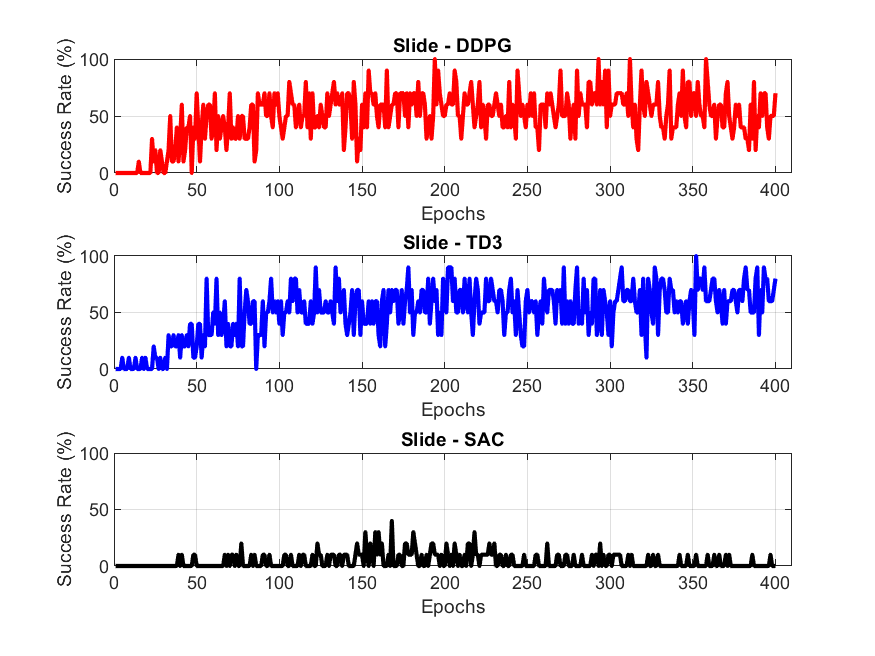}}
	\caption{Slide Comparison}
	\label{fig: slide}
\end{figure}

In this section, the results of DDPG, TD3 and SAC algorithms for four different tasks; \textit{Slide}, \textit{Pick} $\&$ \textit{Place}, \textit{Push} and \textit{Reach}, are given which are procured by testing all algorithms under the same conditions by using the MuJoCo Fetch Robot Manipulator simulation. In order to determine the most ideal model-free off-policy RL algorithm based on the success rate, the epoch time and the efficiency, all three algorithms are trained for 400 epochs in total and 50 episodes for each epoch.\\
In the network models of DDPG algorithm, three fully connected layers with 256 nodes is selected. The structure of actor(policy) and critic(value) networks are exactly the same, except the output layer of the actor network which \textit{tanh} activation function is used. The learning rate, $\alpha$, is accepted as 0.0001 for both actor and critic networks. The discount factor, $\gamma$, for Bellman equation is chosen as 0.98 and polyak, $\rho$, is 0.05 is used to update the target networks. The Gaussian noise is added to the action value during the training process for exploration.\\
The network architecture of TD3 algorithm is also similar to DDPG and consists of actor and critic networks with 3 layers and 256 nodes for each layer. The discount factor, $\gamma$, of the Bellman equation and the learning rate, $\alpha$ of the actor and critic networks are accepted as 0.98 and 0.0001, respectively. In order to update the target networks of the algorithm, the polyak value, $\rho$, is chosen as 0.05 and the Gaussian noise is also added to the action value during training.\\
The neural model of SAC differs from other algorithms and consists of three neural networks, which are actor, critic and q-value. The neural network of SAC has two layers with 256 nodes as recommended. As DDPG and TD3, the learning rate of all networks, $\alpha$, is 0.0001, the discount factor, $\gamma$, of Bellman equation is 0.98 and the polyak, $\rho$, is 0.05. The output layer of actor networks is parametrized with \textit{tanh} function.\\
Figure \ref{fig: slide} presents the general comparison of DDPG, TD3 and SAC algorithms on \textit{Slide} task which is the most challenging task in Fetch simulation environment. SAC algorithm fails to solve the task successfully in 400 epochs but, unlike SAC algorithm, DDPG and TD3 solve the task over 50 percent. Although DDPG has high results in the first 50 epoch, TD3 shows better and higher results at the end of the training. TD3 also has less frequently fluctuation on success rate over the epochs. Overall TD3 algorithm has better results on solving \textit{Slide} task.

\begin{figure}[h]
	\centering
	\framebox{\includegraphics[scale=0.57]{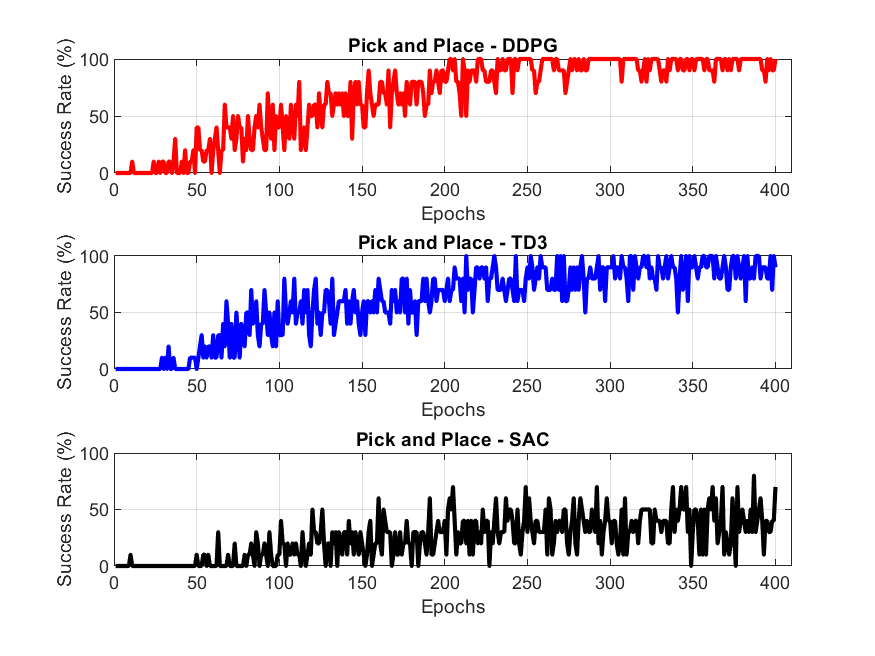}}
	\caption{Pick and Place Comparison}
	\label{fig: pick_place}
\end{figure}

The comparison of next challenging task, \textit{Pick and Place}, is illustrated in Figure \ref{fig: pick_place}. In this task, DDPG shows the highest result, almost 100 percent with some fluctuation after 250 epochs. The incremental slope of success rate is also better in DDPG than others. TD3 finished the task with high rate of success as well, which is around 90 percent, but the learning process is a little bit unstable even after 300 epochs. On the other hand, SAC algorithm shows the lowest score on this task, about 40 percent success rate but its slope exhibits a rising trend. In total, DDPG has better and faster result on this task.

\begin{figure}[h]
	\centering
	\framebox{\includegraphics[scale=0.57]{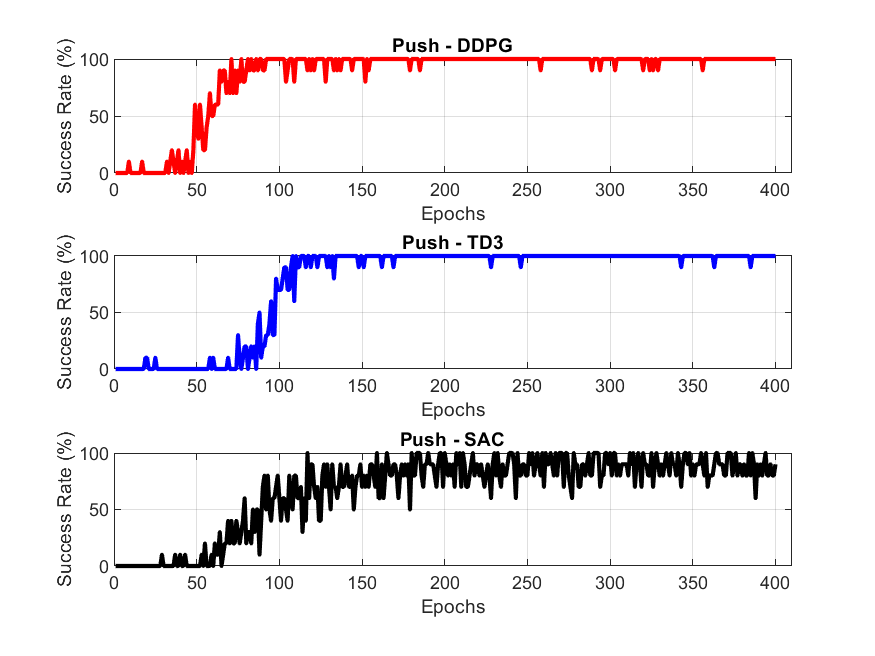}}
	\caption{Push Comparison}
	\label{fig: push}
\end{figure}

In Figure \ref{fig: push}, the results of three off-policy algorithms on \textit{Push} are given. All algorithms successfully solve the task and show high results over 80 percent. DDPG has evolved to 100 percent success rate after $50^{th}$ epochs but TD3 shows the same characteristic after around $80^{th}$ epochs. Despite the fact that DDPG has reached the maximum leve of success at early stage, the total result of TD3 is more stable. Unlike DDPG and TD3, SAC algorithm has highly fluctuating character. Shortly, in terms of speed, DDPG has better result but in terms of stability and success, TD3 can be chosen for \textit{Push} task.

\begin{figure}[thpb]
	\centering
	\framebox{\includegraphics[scale=0.57]{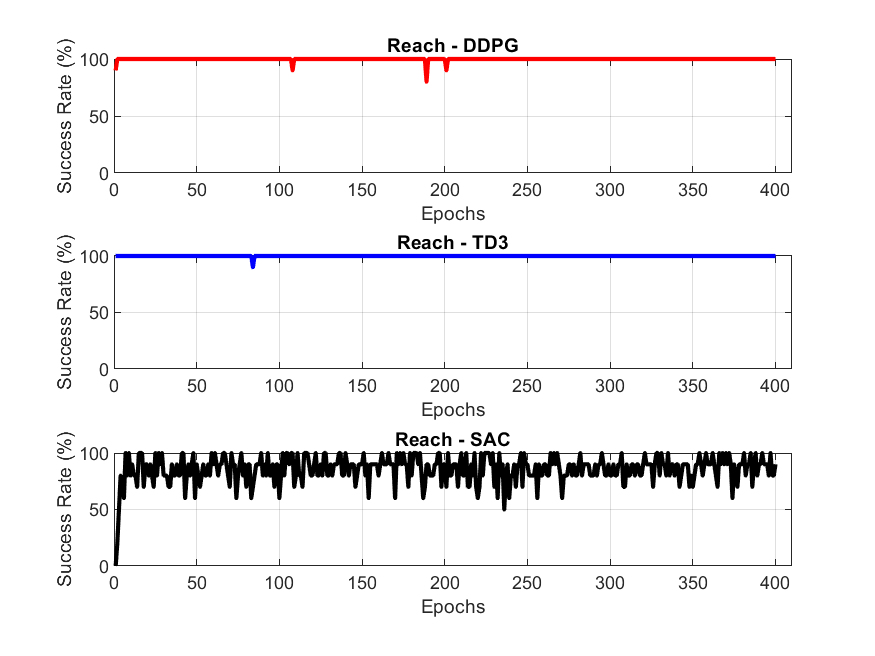}}
	\caption{Reach Comparison}
	\label{fig: reach}
\end{figure}

\textit{Reach} task is the easiest task among the others and the comparison between the algorithms is illustrated in Figure \ref{fig: reach}. DDPG and TD3 achieved 100 percent from the first epoch but SAC algorithm fluctuates around 80 percent during all training. TD2 shows more stable structure from the beginning compared to DDPG and SAC. Thus, TD3 can be chosen for \textit{Reach} task for its stability.

Table \ref{time_table} shows the total time for DDPG, TD3 and SAC algorihtms to finish 400 epochs for \textit{Slide}, \textit{Pick} $\&$ \textit{Place}, \textit{Push} and \textit{Reach} tasks. Here, it can be easily seen that DDPG is the fastest algorithm while SAC showing the slowest time result. The total training time of TD3 is considerably close to the result of DDPG.

\begin{table}[h]
	\caption{The time comparison of algorithms}
	\label{time_table}
	\begin{center}
		\begin{tabular}{|c||c|c|c|c|}
			\hline
			& Slide & Pick$\&$Place & Push & Reach \\
			\hline 
			DDPG & 3h 13m 41s & 3h 11m 55s & 3h 10m 46s & 2h 56m 23s \\
			\hline
			TD3 & 3h 37m 38s & 3h 25m 56s & 3h 31m 14s & 3h 18m 32s	\\
			\hline
			SAC & 4h 5m 38s & 3h 55m 47s & 4h 1m 2s & 3h 46m 20s \\
			\hline
		\end{tabular}
	\end{center}
\end{table}

\begin{figure}
	\centering
	\begin{subfigure}[b]{0.3\textwidth}
		\centering
		\includegraphics[width=\textwidth]{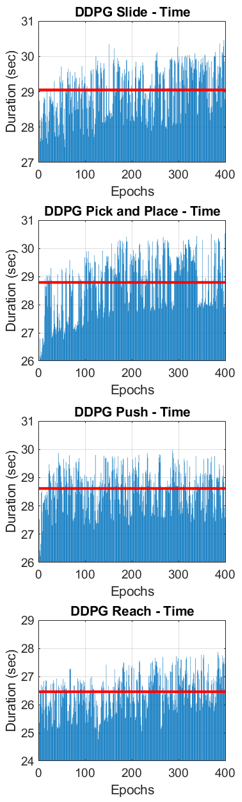}
		\caption{DDPG}
		\label{fig: ddpg_time}
	\end{subfigure}
	\hfill
	\begin{subfigure}[b]{0.3\textwidth}
		\centering
		\includegraphics[width=\textwidth]{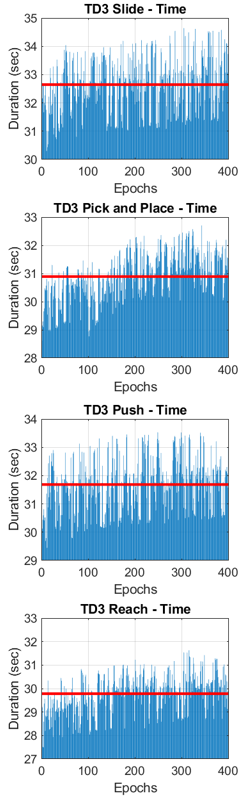}
		\caption{TD3}
		\label{fig: td3}
	\end{subfigure}
	\hfill
	\begin{subfigure}[b]{0.3\textwidth}
		\centering
		\includegraphics[width=\textwidth]{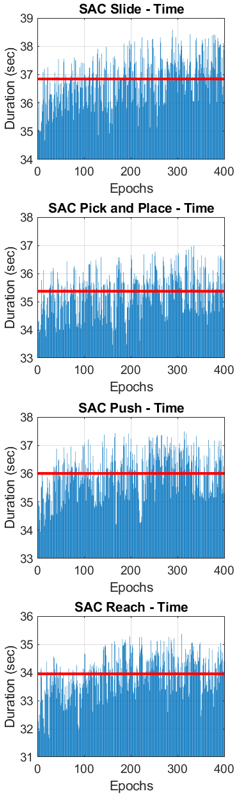}
		\caption{SAC}
		\label{fig: sac}
	\end{subfigure}
	\caption{Time complexity of the applied off-policy algorithms.}
	\label{fig:three_graphs}
\end{figure}

Consequently, every algorithm has different efficiency for each task. In order to solve the \textit{Slide} task, TD3 algorithm shows better result than DDPG and SAC, however, DDPG has the best success rate for \textit{Pick} $\&$ \textit{Place} task. These two tasks are the most challenging tasks because control and stability should be accurate enough in order to achieve the goal. While solving the relatively easier tasks \textit{Push} and \textit{Reach}, TD3 also shows better and stable results than other algorithms. SAC algorithm fails to solve the first two challenging tasks but gives good results for the easier tasks. On the other hand, DDPG finished 400 epochs faster than TD3 and SAC. In short, in order to control a robotic manipulator for different tasks, TD3 shows the best and stable results, but in terms of time, DDPG can be chosen to achieve the task.
Figure \ref{fig:three_graphs} illustrates the episode times of all tasks for the Fetch Robotic. As an instance, considering DDPG, as it is seen from figure \ref{fig: ddpg_time}, for the sliding task, it takes 29 seconds to finish a training epoch (overall success rate is around 50 percent). For Pick and Place task the duration of the training is almost around 29 seconds. Push task is relatively easier than first two tasks with a duration of almost 28 seconds for each epoch. The simplest task, Reach, immediately achieved 100 percent success with small amount of fluctuation and normally takes less time than other tasks with almost 26 seconds.

\section{Conclusion}\label{sec: conclusions}
The study specifically focused on off-policy reinforcement learning algorithms, such as DDPG, TD3, and SAC, and their effectiveness in solving complex control tasks in robotic manipulators. We found that these algorithms can learn to take actions in order to maximize a reward signal, which is essential for tasks such as reaching, picking, placing, and sliding objects. However, the performance and reaction of different algorithms can vary depending on the specific task, which highlights the importance of selecting the most appropriate algorithm for a given task.

In addition to their effectiveness in solving complex control tasks, off-policy reinforcement learning algorithms have the advantage of being model-free, which means they do not require a detailed model of the environment or the system being controlled. This can make them more flexible and adaptable to a wider range of tasks and environments. However, the training process for these algorithms can be time-consuming and require significant computational resources, which emphasizes the need to carefully select the most appropriate algorithm for a given task.

In our paper, we applied DDPG, TD3 and SAC algorithms to train 7-DoF Fetch robotic manipulator, which is based on MuJoCo simulator, in order to achieve four different tasks. After training 400 epochs and 50 episodes for each epoch, DDPG and TD3 algorithms successfully solve the most challenging task, Slide, but SAC algorithm fails on this. All algorithms show successful results on Pick $\&$ Place task; however, SAC algorithm still achieves lower success rate than others. Push and Reach, which are the relatively easier tasks, are solved over 90 percent success rate by all algorithms. While examining all the results of the algorithms, TD3 shows better and stable result for all tasks. On the other hand, in terms of time, DDPG has higher successful result but has a small difference with respect to TD3. Shortly, TD3 is the most successful algorithm among the off-policy model free algorithms in order to solve robotic manipulator tasks according to this study. \\
On the other hand, in order to claim an algorithm as the best option, it needs to be tested on a real-life system. As a future work, DDPG, TD3 and SAC algorithms will be tested on a 7-DoF robotic manipulator, either Fetch or Kuka, in order to physically implement the given tasks. 



\hyphenpenalty=10000 

\bibliographystyle{cas-model2-names}			

\bibliography{refs}

\begin{thebibliography}{47}
\expandafter\ifx\csname natexlab\endcsname\relax\def\natexlab#1{#1}\fi
\providecommand{\url}[1]{\texttt{#1}}
\providecommand{\href}[2]{#2}
\providecommand{\path}[1]{#1}
\providecommand{\DOIprefix}{doi:}
\providecommand{\ArXivprefix}{arXiv:}
\providecommand{\URLprefix}{URL: }
\providecommand{\Pubmedprefix}{pmid:}
\providecommand{\doi}[1]{\href{http://dx.doi.org/#1}{\path{#1}}}
\providecommand{\Pubmed}[1]{\href{pmid:#1}{\path{#1}}}
\providecommand{\bibinfo}[2]{#2}
\ifx\xfnm\relax \def\xfnm[#1]{\unskip,\space#1}\fi
\bibitem[{Akbulut et~al.(2020)Akbulut, Oztop, Seker, Xue, Tekden and
  Ugur}]{emreugur20}
\bibinfo{author}{Akbulut, M.T.}, \bibinfo{author}{Oztop, E.},
  \bibinfo{author}{Seker, M.Y.}, \bibinfo{author}{Xue, H.},
  \bibinfo{author}{Tekden, A.E.}, \bibinfo{author}{Ugur, E.},
  \bibinfo{year}{2020}.
\newblock \bibinfo{title}{Acnmp: Skill transfer and task extrapolation through
  learning from demonstration and reinforcement learning via representation
  sharing}.
\newblock \DOIprefix\doi{10.48550/ARXIV.2003.11334}.
\bibitem[{Aumjaud et~al.(2020)Aumjaud, McAuliffe, Lera and
  Cardiff}]{Aumjaud2020}
\bibinfo{author}{Aumjaud, P.}, \bibinfo{author}{McAuliffe, D.},
  \bibinfo{author}{Lera, F.J.R.}, \bibinfo{author}{Cardiff, P.},
  \bibinfo{year}{2020}.
\newblock \bibinfo{title}{Reinforcement learning experiments and benchmark for
  solving robotic reaching tasks}.
\newblock \bibinfo{journal}{ArXiv} \bibinfo{volume}{abs/2011.05782}.
\bibitem[{Belousov et~al.(2021)Belousov, H., Klink, Parisi and
  Peters}]{BelAbdKliParPet21Book}
\bibinfo{author}{Belousov, B.}, \bibinfo{author}{H., A.},
  \bibinfo{author}{Klink, P.}, \bibinfo{author}{Parisi, S.},
  \bibinfo{author}{Peters, J.}, \bibinfo{year}{2021}.
\newblock \bibinfo{title}{Reinforcement Learning Algorithms: Analysis and
  Applications}. volume \bibinfo{volume}{883} of
  \textit{\bibinfo{series}{Studies in Computational Intelligence}}.
\newblock \bibinfo{publisher}{Springer International Publishing}.
\newblock \DOIprefix\doi{10.1007/978-3-030-41188-6}.
\bibitem[{Bertino et~al.(2022)Bertino, Naseradinmousavi and Krstić}]{peiman22}
\bibinfo{author}{Bertino, A.}, \bibinfo{author}{Naseradinmousavi, P.},
  \bibinfo{author}{Krstić, M.}, \bibinfo{year}{2022}.
\newblock \bibinfo{title}{Delay-adaptive control of a 7-dof robot manipulator:
  Design and experiments}.
\newblock \bibinfo{journal}{IEEE Transactions on Control Systems Technology}
  \bibinfo{volume}{30}, \bibinfo{pages}{2506--2521}.
\newblock \DOIprefix\doi{10.1109/TCST.2022.3152363}.
\bibitem[{Beyret et~al.(2019)Beyret, Shafti and Faisal}]{8968488}
\bibinfo{author}{Beyret, B.}, \bibinfo{author}{Shafti, A.},
  \bibinfo{author}{Faisal, A.A.}, \bibinfo{year}{2019}.
\newblock \bibinfo{title}{Dot-to-dot: Explainable hierarchical reinforcement
  learning for robotic manipulation}, in: \bibinfo{booktitle}{2019 IEEE/RSJ
  International Conference on Intelligent Robots and Systems (IROS)}, pp.
  \bibinfo{pages}{5014--5019}.
\newblock \DOIprefix\doi{10.1109/IROS40897.2019.8968488}.
\bibitem[{Büchler et~al.(2022)Büchler, Guist, Calandra, Berenz, Schölkopf
  and Peters}]{peterstennis22}
\bibinfo{author}{Büchler, D.}, \bibinfo{author}{Guist, S.},
  \bibinfo{author}{Calandra, R.}, \bibinfo{author}{Berenz, V.},
  \bibinfo{author}{Schölkopf, B.}, \bibinfo{author}{Peters, J.},
  \bibinfo{year}{2022}.
\newblock \bibinfo{title}{Learning to play table tennis from scratch using
  muscular robots.}
\newblock \bibinfo{journal}{IEEE Transactions on Robotics} ,
  \bibinfo{pages}{1--11}\DOIprefix\doi{10.1109/TRO.2022.3176207}.
\bibitem[{Casas(2017)}]{trafficDDPG}
\bibinfo{author}{Casas, N.}, \bibinfo{year}{2017}.
\newblock \bibinfo{title}{Deep deterministic policy gradient for urban traffic
  light control}.
\newblock \DOIprefix\doi{10.48550/ARXIV.1703.09035}.
\bibitem[{Dankwa and Zheng(2020)}]{Dankwa19}
\bibinfo{author}{Dankwa, S.}, \bibinfo{author}{Zheng, W.},
  \bibinfo{year}{2020}.
\newblock \bibinfo{title}{Twin-delayed ddpg: A deep reinforcement learning
  technique to model a continuous movement of an intelligent robot agent}, in:
  \bibinfo{booktitle}{Proceedings of the 3rd International Conference on
  Vision, Image and Signal Processing}, \bibinfo{publisher}{Association for
  Computing Machinery}, \bibinfo{address}{New York, NY, USA}.
\newblock \DOIprefix\doi{10.1145/3387168.3387199}.
\bibitem[{Fujimoto et~al.(2018a)Fujimoto, van Hoof and Meger}]{twin18}
\bibinfo{author}{Fujimoto, S.}, \bibinfo{author}{van Hoof, H.},
  \bibinfo{author}{Meger, D.}, \bibinfo{year}{2018}a.
\newblock \bibinfo{title}{Addressing function approximation error in
  actor-critic methods}.
\newblock \DOIprefix\doi{10.48550/ARXIV.1802.09477}.
\bibitem[{Fujimoto et~al.(2018b)Fujimoto, van Hoof and Meger}]{c5}
\bibinfo{author}{Fujimoto, S.}, \bibinfo{author}{van Hoof, H.},
  \bibinfo{author}{Meger, D.}, \bibinfo{year}{2018}b.
\newblock \bibinfo{title}{Addressing function approximation error in
  actor-critic methods.}
\newblock \bibinfo{journal}{CoRR} \bibinfo{volume}{abs/1802.09477}.
\bibitem[{Gao et~al.(2020)Gao, Graesser, Choromanski, Song, Lazic, Sanketi,
  Sindhwani and Jaitly}]{tennis20IROS}
\bibinfo{author}{Gao, W.}, \bibinfo{author}{Graesser, L.},
  \bibinfo{author}{Choromanski, K.}, \bibinfo{author}{Song, X.},
  \bibinfo{author}{Lazic, N.}, \bibinfo{author}{Sanketi, P.},
  \bibinfo{author}{Sindhwani, V.}, \bibinfo{author}{Jaitly, N.},
  \bibinfo{year}{2020}.
\newblock \bibinfo{title}{Robotic table tennis with model-free reinforcement
  learning}, in: \bibinfo{booktitle}{2020 IEEE/RSJ International Conference on
  Intelligent Robots and Systems (IROS)}, pp. \bibinfo{pages}{5556--5563}.
\newblock \DOIprefix\doi{10.1109/IROS45743.2020.9341191}.
\bibitem[{García and Shafie(2020)}]{humanoid2020}
\bibinfo{author}{García, J.}, \bibinfo{author}{Shafie, D.},
  \bibinfo{year}{2020}.
\newblock \bibinfo{title}{Teaching a humanoid robot to walk faster through safe
  reinforcement learning}.
\newblock \bibinfo{journal}{Engineering Applications of Artificial
  Intelligence} \bibinfo{volume}{88}, \bibinfo{pages}{103360}.
\newblock \DOIprefix\doi{https://doi.org/10.1016/j.engappai.2019.103360}.
\bibitem[{Haarnoja et~al.(2018)Haarnoja, Zhou, Abbeel and Levine}]{SAC18}
\bibinfo{author}{Haarnoja, T.}, \bibinfo{author}{Zhou, A.},
  \bibinfo{author}{Abbeel, P.}, \bibinfo{author}{Levine, S.},
  \bibinfo{year}{2018}.
\newblock \bibinfo{title}{Soft actor-critic: Off-policy maximum entropy deep
  reinforcement learning with a stochastic actor}.
\newblock \DOIprefix\doi{10.48550/ARXIV.1801.01290}.
\bibitem[{Hafner et~al.(2019)Hafner, Lillicrap, Ba and Norouzi}]{Hafner1}
\bibinfo{author}{Hafner, D.}, \bibinfo{author}{Lillicrap, T.},
  \bibinfo{author}{Ba, J.}, \bibinfo{author}{Norouzi, M.},
  \bibinfo{year}{2019}.
\newblock \bibinfo{title}{Dream to control: Learning behaviors by latent
  imagination}.
\newblock \DOIprefix\doi{10.48550/ARXIV.1912.01603}.
\bibitem[{Hafner et~al.(2018)Hafner, Lillicrap, Fischer, Villegas, Ha, Lee and
  Davidson}]{Hafner2}
\bibinfo{author}{Hafner, D.}, \bibinfo{author}{Lillicrap, T.},
  \bibinfo{author}{Fischer, I.}, \bibinfo{author}{Villegas, R.},
  \bibinfo{author}{Ha, D.}, \bibinfo{author}{Lee, H.},
  \bibinfo{author}{Davidson, J.}, \bibinfo{year}{2018}.
\newblock \bibinfo{title}{Learning latent dynamics for planning from pixels}.
\newblock \DOIprefix\doi{10.48550/ARXIV.1811.04551}.
\bibitem[{Imani et~al.(2018)Imani, Graves and White}]{offpolicy18imani}
\bibinfo{author}{Imani, E.}, \bibinfo{author}{Graves, E.},
  \bibinfo{author}{White, M.}, \bibinfo{year}{2018}.
\newblock \bibinfo{title}{An off-policy policy gradient theorem using emphatic
  weightings}, in: \bibinfo{booktitle}{NeurIPS}.
\bibitem[{Kiran et~al.(2022a)Kiran, Sobh, Talpaert, Mannion, Sallab, Yogamani
  and Pérez}]{driving}
\bibinfo{author}{Kiran, B.R.}, \bibinfo{author}{Sobh, I.},
  \bibinfo{author}{Talpaert, V.}, \bibinfo{author}{Mannion, P.},
  \bibinfo{author}{Sallab, A.A.A.}, \bibinfo{author}{Yogamani, S.},
  \bibinfo{author}{Pérez, P.}, \bibinfo{year}{2022}a.
\newblock \bibinfo{title}{Deep reinforcement learning for autonomous driving: A
  survey}.
\newblock \bibinfo{journal}{IEEE Transactions on Intelligent Transportation
  Systems} \bibinfo{volume}{23}, \bibinfo{pages}{4909--4926}.
\newblock \DOIprefix\doi{10.1109/TITS.2021.3054625}.
\bibitem[{Kiran et~al.(2022b)Kiran, Sobh, Talpaert, Mannion, Sallab, Yogamani
  and Pérez}]{survey2022deep}
\bibinfo{author}{Kiran, B.R.}, \bibinfo{author}{Sobh, I.},
  \bibinfo{author}{Talpaert, V.}, \bibinfo{author}{Mannion, P.},
  \bibinfo{author}{Sallab, A.A.A.}, \bibinfo{author}{Yogamani, S.},
  \bibinfo{author}{Pérez, P.}, \bibinfo{year}{2022}b.
\newblock \bibinfo{title}{Deep reinforcement learning for autonomous driving: A
  survey}.
\newblock \bibinfo{journal}{IEEE Transactions on Intelligent Transportation
  Systems} \bibinfo{volume}{23}, \bibinfo{pages}{4909--4926}.
\newblock \DOIprefix\doi{10.1109/TITS.2021.3054625}.
\bibitem[{Kosaka(2019)}]{c17}
\bibinfo{author}{Kosaka, N.}, \bibinfo{year}{2019}.
\newblock \bibinfo{title}{Has it explored enough?}
\newblock \bibinfo{type}{Master's thesis}. Royal Holloway University of London.
  \bibinfo{address}{London}.
\bibitem[{Kumar et~al.(2018)Kumar, Paul and Omkar}]{kumar18}
\bibinfo{author}{Kumar, A.}, \bibinfo{author}{Paul, N.},
  \bibinfo{author}{Omkar, S.N.}, \bibinfo{year}{2018}.
\newblock \bibinfo{title}{Bipedal walking robot using deep deterministic policy
  gradient}.
\newblock \bibinfo{journal}{CoRR} \bibinfo{volume}{abs/1807.05924}.
\newblock \href{http://arxiv.org/abs/1807.05924}{\tt arXiv:1807.05924}.
\bibitem[{Li et~al.(2021)Li, Shi, Chen, Gu and Chi}]{modelfree21_1}
\bibinfo{author}{Li, G.}, \bibinfo{author}{Shi, L.}, \bibinfo{author}{Chen,
  Y.}, \bibinfo{author}{Gu, Y.}, \bibinfo{author}{Chi, Y.},
  \bibinfo{year}{2021}.
\newblock \bibinfo{title}{Breaking the sample complexity barrier to
  regret-optimal model-free reinforcement learning}, in:
  \bibinfo{editor}{Ranzato, M.}, \bibinfo{editor}{Beygelzimer, A.},
  \bibinfo{editor}{Dauphin, Y.}, \bibinfo{editor}{Liang, P.},
  \bibinfo{editor}{Vaughan, J.W.} (Eds.), \bibinfo{booktitle}{Advances in
  Neural Information Processing Systems}, \bibinfo{publisher}{Curran
  Associates, Inc.}. pp. \bibinfo{pages}{17762--17776}.
\bibitem[{Lillicrap et~al.(2015)Lillicrap, Hunt, Pritzel, Heess, Erez, Tassa,
  Silver and Wierstra}]{ddpg15}
\bibinfo{author}{Lillicrap, T.P.}, \bibinfo{author}{Hunt, J.J.},
  \bibinfo{author}{Pritzel, A.}, \bibinfo{author}{Heess, N.},
  \bibinfo{author}{Erez, T.}, \bibinfo{author}{Tassa, Y.},
  \bibinfo{author}{Silver, D.}, \bibinfo{author}{Wierstra, D.},
  \bibinfo{year}{2015}.
\newblock \bibinfo{title}{Continuous control with deep reinforcement learning}.
\newblock \DOIprefix\doi{10.48550/ARXIV.1509.02971}.
\bibitem[{Lindner and Milecki(2022)}]{obstacleavoid22}
\bibinfo{author}{Lindner, T.}, \bibinfo{author}{Milecki, A.},
  \bibinfo{year}{2022}.
\newblock \bibinfo{title}{Reinforcement learning-based algorithm to avoid
  obstacles by the anthropomorphic robotic arm}.
\newblock \bibinfo{journal}{Applied Sciences} \bibinfo{volume}{12}.
\newblock \DOIprefix\doi{10.3390/app12136629}.
\bibitem[{Liu et~al.(2022)Liu, Zhang, Tateo, Jauhri, Hu, Peters and
  Chalvatzaki}]{peters2022RL}
\bibinfo{author}{Liu, P.}, \bibinfo{author}{Zhang, K.}, \bibinfo{author}{Tateo,
  D.}, \bibinfo{author}{Jauhri, S.}, \bibinfo{author}{Hu, Z.},
  \bibinfo{author}{Peters, J.}, \bibinfo{author}{Chalvatzaki, G.},
  \bibinfo{year}{2022}.
\newblock \bibinfo{title}{Safe reinforcement learning of dynamic
  high-dimensional robotic tasks: navigation, manipulation, interaction}.
\newblock \DOIprefix\doi{10.48550/ARXIV.2209.13308}.
\bibitem[{Luo et~al.(2022)Luo, Xu, Lai, Chen, Zhang and Yu}]{MBRL22survey}
\bibinfo{author}{Luo, F.M.}, \bibinfo{author}{Xu, T.}, \bibinfo{author}{Lai,
  H.}, \bibinfo{author}{Chen, X.H.}, \bibinfo{author}{Zhang, W.},
  \bibinfo{author}{Yu, Y.}, \bibinfo{year}{2022}.
\newblock \bibinfo{title}{A survey on model-based reinforcement learning}.
\newblock \DOIprefix\doi{10.48550/ARXIV.2206.09328}.
\bibitem[{Lv et~al.(2022)Lv, Feng, Zhang, Zhao, Shao and Lu}]{pancake22}
\bibinfo{author}{Lv, J.}, \bibinfo{author}{Feng, Y.}, \bibinfo{author}{Zhang,
  C.}, \bibinfo{author}{Zhao, S.}, \bibinfo{author}{Shao, L.},
  \bibinfo{author}{Lu, C.}, \bibinfo{year}{2022}.
\newblock \bibinfo{title}{Sam-rl: Sensing-aware model-based reinforcement
  learning via differentiable physics-based simulation and rendering}.
\newblock \DOIprefix\doi{10.48550/ARXIV.2210.15185}.
\bibitem[{Mnih et~al.(2016)Mnih, Badia, Mirza, Graves, Lillicrap, Harley,
  Silver and Kavukcuoglu}]{Mniha2c16}
\bibinfo{author}{Mnih, V.}, \bibinfo{author}{Badia, A.P.},
  \bibinfo{author}{Mirza, M.}, \bibinfo{author}{Graves, A.},
  \bibinfo{author}{Lillicrap, T.P.}, \bibinfo{author}{Harley, T.},
  \bibinfo{author}{Silver, D.}, \bibinfo{author}{Kavukcuoglu, K.},
  \bibinfo{year}{2016}.
\newblock \bibinfo{title}{Asynchronous methods for deep reinforcement learning}
  \DOIprefix\doi{10.48550/ARXIV.1602.01783}.
\bibitem[{Mnih et~al.(2015)Mnih, Kavukcuoglu, Silver, Rusu, Veness, Bellemare,
  Graves, Riedmiller, Fidjeland, Ostrovski, Petersen, Beattie, Sadik,
  Antonoglou, King, Kumaran, Wierstra, Legg and Hassabis}]{mnih2015DQN}
\bibinfo{author}{Mnih, V.}, \bibinfo{author}{Kavukcuoglu, K.},
  \bibinfo{author}{Silver, D.}, \bibinfo{author}{Rusu, A.A.},
  \bibinfo{author}{Veness, J.}, \bibinfo{author}{Bellemare, M.G.},
  \bibinfo{author}{Graves, A.}, \bibinfo{author}{Riedmiller, M.},
  \bibinfo{author}{Fidjeland, A.K.}, \bibinfo{author}{Ostrovski, G.},
  \bibinfo{author}{Petersen, S.}, \bibinfo{author}{Beattie, C.},
  \bibinfo{author}{Sadik, A.}, \bibinfo{author}{Antonoglou, I.},
  \bibinfo{author}{King, H.}, \bibinfo{author}{Kumaran, D.},
  \bibinfo{author}{Wierstra, D.}, \bibinfo{author}{Legg, S.},
  \bibinfo{author}{Hassabis, D.}, \bibinfo{year}{2015}.
\newblock \bibinfo{title}{Human-level control through deep reinforcement
  learning}.
\newblock \bibinfo{journal}{Nature} \bibinfo{volume}{518},
  \bibinfo{pages}{529--533}.
\newblock \DOIprefix\doi{10.1038/nature14236}.
\bibitem[{Nair et~al.(2018)Nair, McGrew, Andrychowicz, Zaremba and
  Abbeel}]{8463162}
\bibinfo{author}{Nair, A.}, \bibinfo{author}{McGrew, B.},
  \bibinfo{author}{Andrychowicz, M.}, \bibinfo{author}{Zaremba, W.},
  \bibinfo{author}{Abbeel, P.}, \bibinfo{year}{2018}.
\newblock \bibinfo{title}{Overcoming exploration in reinforcement learning with
  demonstrations}, in: \bibinfo{booktitle}{2018 IEEE International Conference
  on Robotics and Automation (ICRA)}, pp. \bibinfo{pages}{6292--6299}.
\newblock \DOIprefix\doi{10.1109/ICRA.2018.8463162}.
\bibitem[{Nguyen et~al.(2018)Nguyen, La and Deans}]{Nguyan18Fetch}
\bibinfo{author}{Nguyen, H.}, \bibinfo{author}{La, H.M.},
  \bibinfo{author}{Deans, M.}, \bibinfo{year}{2018}.
\newblock \bibinfo{title}{Deep learning with experience ranking convolutional
  neural network for robot manipulator}.
\newblock \DOIprefix\doi{10.48550/ARXIV.1809.05819}.
\bibitem[{{OpenAI} et~al.(2019){OpenAI}, Akkaya, Andrychowicz, Chociej, Litwin,
  McGrew, Petron, Paino, Plappert, Powell, Ribas, Schneider, Tezak, Tworek,
  Welinder, Weng, Yuan, Zaremba and Zhang}]{rubikcube}
\bibinfo{author}{{OpenAI}}, \bibinfo{author}{Akkaya, I.},
  \bibinfo{author}{Andrychowicz, M.}, \bibinfo{author}{Chociej, M.},
  \bibinfo{author}{Litwin, M.}, \bibinfo{author}{McGrew, B.},
  \bibinfo{author}{Petron, A.}, \bibinfo{author}{Paino, A.},
  \bibinfo{author}{Plappert, M.}, \bibinfo{author}{Powell, G.},
  \bibinfo{author}{Ribas, R.}, \bibinfo{author}{Schneider, J.},
  \bibinfo{author}{Tezak, N.}, \bibinfo{author}{Tworek, J.},
  \bibinfo{author}{Welinder, P.}, \bibinfo{author}{Weng, L.},
  \bibinfo{author}{Yuan, Q.}, \bibinfo{author}{Zaremba, W.},
  \bibinfo{author}{Zhang, L.}, \bibinfo{year}{2019}.
\newblock \bibinfo{title}{Solving rubik's cube with a robot hand}.
\newblock \DOIprefix\doi{10.48550/ARXIV.1910.07113}.
\bibitem[{Peters and Schaal(2006)}]{petersPG2006}
\bibinfo{author}{Peters, J.}, \bibinfo{author}{Schaal, S.},
  \bibinfo{year}{2006}.
\newblock \bibinfo{title}{Policy gradient methods for robotics}, in:
  \bibinfo{booktitle}{2006 IEEE/RSJ International Conference on Intelligent
  Robots and Systems}, pp. \bibinfo{pages}{2219--2225}.
\newblock \DOIprefix\doi{10.1109/IROS.2006.282564}.
\bibitem[{Rawlik et~al.(2012)Rawlik, Toussaint and Vijayakumar}]{RawlikTV12}
\bibinfo{author}{Rawlik, K.}, \bibinfo{author}{Toussaint, M.},
  \bibinfo{author}{Vijayakumar, S.}, \bibinfo{year}{2012}.
\newblock \bibinfo{title}{On stochastic optimal control and reinforcement
  learning by approximate inference.}, in: \bibinfo{editor}{Roy, N.},
  \bibinfo{editor}{Newman, P.}, \bibinfo{editor}{Srinivasa, S.S.} (Eds.),
  \bibinfo{booktitle}{Robotics: Science and Systems}.
\bibitem[{Shahid et~al.(2022)Shahid, Piga, Braghin and
  Roveda}]{Shahid2022roboticRL}
\bibinfo{author}{Shahid, A.A.}, \bibinfo{author}{Piga, D.},
  \bibinfo{author}{Braghin, F.}, \bibinfo{author}{Roveda, L.},
  \bibinfo{year}{2022}.
\newblock \bibinfo{title}{Continuous control actions learning and adaptation
  for robotic manipulation through reinforcement learning}.
\newblock \bibinfo{journal}{Auton. Robots} \bibinfo{volume}{46},
  \bibinfo{pages}{483–498}.
\newblock \DOIprefix\doi{10.1007/s10514-022-10034-z}.
\bibitem[{Silver et~al.(2014)Silver, Lever, Heess, Degris, Wierstra and
  Riedmiller}]{Silverconf}
\bibinfo{author}{Silver, D.}, \bibinfo{author}{Lever, G.},
  \bibinfo{author}{Heess, N.}, \bibinfo{author}{Degris, T.},
  \bibinfo{author}{Wierstra, D.}, \bibinfo{author}{Riedmiller, M.A.},
  \bibinfo{year}{2014}.
\newblock \bibinfo{title}{Deterministic policy gradient algorithms.}, in:
  \bibinfo{booktitle}{ICML}, \bibinfo{publisher}{JMLR.org}. pp.
  \bibinfo{pages}{387--395}.
\bibitem[{Steckelmacher et~al.(2019)Steckelmacher, Plisnier, Roijers and
  Nowé}]{c20}
\bibinfo{author}{Steckelmacher, D.}, \bibinfo{author}{Plisnier, H.},
  \bibinfo{author}{Roijers, D.M.}, \bibinfo{author}{Nowé, A.},
  \bibinfo{year}{2019}.
\newblock \bibinfo{title}{Sample-efficient model-free reinforcement learning
  with off-policy critics} \DOIprefix\doi{10.48550/ARXIV.1903.04193}.
\bibitem[{{Tavakol Aghaei} et~al.(2022){Tavakol Aghaei}, Ağababaoğlu,
  Yıldırım and Onat}]{TAVAKOLAGHAEI2021}
\bibinfo{author}{{Tavakol Aghaei}, V.}, \bibinfo{author}{Ağababaoğlu, A.},
  \bibinfo{author}{Yıldırım, S.}, \bibinfo{author}{Onat, A.},
  \bibinfo{year}{2022}.
\newblock \bibinfo{title}{A real-world application of markov chain monte carlo
  method for bayesian trajectory control of a robotic manipulator}.
\newblock \bibinfo{journal}{ISA Transactions} \bibinfo{volume}{125},
  \bibinfo{pages}{580--590}.
\newblock \DOIprefix\doi{https://doi.org/10.1016/j.isatra.2021.06.010}.
\bibitem[{Todorov(2008)}]{Todorov08}
\bibinfo{author}{Todorov, E.}, \bibinfo{year}{2008}.
\newblock \bibinfo{title}{General duality between optimal control and
  estimation.}, in: \bibinfo{booktitle}{CDC}, \bibinfo{publisher}{IEEE}. pp.
  \bibinfo{pages}{4286--4292}.
\bibitem[{Todorov et~al.(2012)Todorov, Erez and Tassa}]{c18}
\bibinfo{author}{Todorov, E.}, \bibinfo{author}{Erez, T.},
  \bibinfo{author}{Tassa, Y.}, \bibinfo{year}{2012}.
\newblock \bibinfo{title}{Mujoco: A physics engine for model-based control.},
  in: \bibinfo{booktitle}{IROS}, \bibinfo{publisher}{IEEE}. pp.
  \bibinfo{pages}{5026--5033}.
\bibitem[{Toussaint(2009)}]{Toussaint09}
\bibinfo{author}{Toussaint, M.}, \bibinfo{year}{2009}.
\newblock \bibinfo{title}{Robot trajectory optimization using approximate
  inference.}, in: \bibinfo{editor}{Danyluk, A.P.}, \bibinfo{editor}{Bottou,
  L.}, \bibinfo{editor}{Littman, M.L.} (Eds.), \bibinfo{booktitle}{ICML},
  \bibinfo{publisher}{ACM}. pp. \bibinfo{pages}{1049--1056}.
\bibitem[{Tu et~al.(2021)Tu, Fei and Deng}]{flip2021}
\bibinfo{author}{Tu, Z.}, \bibinfo{author}{Fei, F.}, \bibinfo{author}{Deng,
  X.}, \bibinfo{year}{2021}.
\newblock \bibinfo{title}{Bio-inspired rapid escape and tight body flip on an
  at-scale flapping wing hummingbird robot via reinforcement learning}.
\newblock \bibinfo{journal}{IEEE Transactions on Robotics}
  \bibinfo{volume}{37}, \bibinfo{pages}{1742--1751}.
\newblock \DOIprefix\doi{10.1109/TRO.2021.3064882}.
\bibitem[{Tuyen and Chung(2017)}]{TuyenC17}
\bibinfo{author}{Tuyen, L.P.}, \bibinfo{author}{Chung, T.},
  \bibinfo{year}{2017}.
\newblock \bibinfo{title}{Controlling bicycle using deep deterministic policy
  gradient algorithm.}, in: \bibinfo{booktitle}{URAI},
  \bibinfo{publisher}{IEEE}. pp. \bibinfo{pages}{413--417}.
\bibitem[{Woo et~al.(2020)Woo, Wu, Park and Roh}]{woo20}
\bibinfo{author}{Woo, J.H.}, \bibinfo{author}{Wu, L.}, \bibinfo{author}{Park,
  J.B.}, \bibinfo{author}{Roh, J.H.}, \bibinfo{year}{2020}.
\newblock \bibinfo{title}{Real-time optimal power flow using twin delayed deep
  deterministic policy gradient algorithm}.
\newblock \bibinfo{journal}{IEEE Access} \bibinfo{volume}{8},
  \bibinfo{pages}{213611--213618}.
\newblock \DOIprefix\doi{10.1109/ACCESS.2020.3041007}.
\bibitem[{Wu et~al.(2017)Wu, Mansimov, Liao, Grosse and Ba}]{wu17}
\bibinfo{author}{Wu, Y.}, \bibinfo{author}{Mansimov, E.},
  \bibinfo{author}{Liao, S.}, \bibinfo{author}{Grosse, R.},
  \bibinfo{author}{Ba, J.}, \bibinfo{year}{2017}.
\newblock \bibinfo{title}{Scalable trust-region method for deep reinforcement
  learning using kronecker-factored approximation}.
\newblock \DOIprefix\doi{10.48550/ARXIV.1708.05144}.
\bibitem[{Xie et~al.(2022)Xie, Dong, Zhao and Karcanias}]{DDPGwind}
\bibinfo{author}{Xie, J.}, \bibinfo{author}{Dong, H.}, \bibinfo{author}{Zhao,
  X.}, \bibinfo{author}{Karcanias, A.}, \bibinfo{year}{2022}.
\newblock \bibinfo{title}{Wind farm power generation control via
  double-network-based deep reinforcement learning}.
\newblock \bibinfo{journal}{IEEE Transactions on Industrial Informatics}
  \bibinfo{volume}{18}, \bibinfo{pages}{2321--2330}.
\newblock \DOIprefix\doi{10.1109/TII.2021.3095563}.
\bibitem[{Yu et~al.(2021)Yu, Tang, Huang, Zhang, So and Wong}]{UAV21}
\bibinfo{author}{Yu, Y.}, \bibinfo{author}{Tang, J.}, \bibinfo{author}{Huang,
  J.}, \bibinfo{author}{Zhang, X.}, \bibinfo{author}{So, D.K.C.},
  \bibinfo{author}{Wong, K.K.}, \bibinfo{year}{2021}.
\newblock \bibinfo{title}{Multi-objective optimization for uav-assisted
  wireless powered iot networks based on extended ddpg algorithm}.
\newblock \bibinfo{journal}{IEEE Transactions on Communications}
  \bibinfo{volume}{69}, \bibinfo{pages}{6361--6374}.
\newblock \DOIprefix\doi{10.1109/TCOMM.2021.3089476}.
\bibitem[{Ziebart et~al.(2008)Ziebart, Maas, Bagnell and Dey}]{ZiebartMBD08}
\bibinfo{author}{Ziebart, B.D.}, \bibinfo{author}{Maas, A.L.},
  \bibinfo{author}{Bagnell, J.A.}, \bibinfo{author}{Dey, A.K.},
  \bibinfo{year}{2008}.
\newblock \bibinfo{title}{Maximum entropy inverse reinforcement learning.}, in:
  \bibinfo{editor}{Fox, D.}, \bibinfo{editor}{Gomes, C.P.} (Eds.),
  \bibinfo{booktitle}{AAAI}, \bibinfo{publisher}{AAAI Press}. pp.
  \bibinfo{pages}{1433--1438}.

\end{thebibliography}

\vskip6pt
\end{document}